# Real-Time Pothole Detection Using Deep Learning

Anas Al-Shaghouri, Rami Alkhatib, Samir Berjaoui

*Abstract* — Roads are connecting lines between different places and are used in our daily life. Roads' periodic maintenance keeps them safe and functional. However, asphalt pavement distresses cause potholes which may increase the number of accidents. Detecting and reporting the existence of potholes to responsible departments can save the roads from getting worse. This study deployed and tested different deep learning architectures to detect the presence of potholes. First, several images of potholes are captured by a cellphone mounted on the car windshield. Then pothole images downloaded from the internet increased the size and the variability of our database (1087 images with more than 2000 potholes). Second, various object detection algorithms are employed to detect potholes in the road images. Real-time Deep Learning algorithms with several configurations like SSD-TensorFlow, YOLOv3-Darknet53, and YOLOv4-CSPDarknet53 are used to compare their performances on pothole detection. YOLOv4 achieved a high recall of 81%, high precision of 85% and 85.39% mean Average Precision (mAP). The speed of YOLOv4 processing is recorded around 20 frames per second (FPS) at an image resolution of 832*832. Furthermore, the proposed system detected potholes at distances reaching a hundred meters. Compared with other state-of-the-art methods, our result demonstrated superior performance in real-time. The proposed method can help in reporting road potholes to government agencies, increasing the safety of drivers by detecting potholes time ahead, and improving the performance of self-driving cars to ensure safe trips for passengers in the future.

Impact Statement — Potholes cost governments and citizens billions of dollars yearly and lead to 425,000 deaths each year. Reporting potholes to the responsible governmental agencies at an early stage will save a lot of lives. Furthermore, having a real-time pothole detection system implemented in vehicles can alert drivers about the presence of potholes ahead of time. This would help drivers to avoid potholes and protect their vehicles from serious damage. In the last few years, automotive companies have been investing considerably in implementing pothole detection systems to newly released car models in the market, especially in autonomous cars. The real-time detection system using Deep Learning, proposed in this paper, has a superior performance compared to other previously proposed systems in the literature.

A. A. is with the Mechanical and Mechatronics Engineering Department, Rafik Hairri University, Mechref- Chouf2010, Lebanon. (correspondence e-mail: alshaghouriam@rhu.edu.lb)
R. A. is with the Mechanical and Mechatronics Engineering Department, Rafik Hairri University, Mechref- Chouf2010, Lebanon. (correspondence e-mail: khatibrh@rhu.edu.lb)
S. B. is with the College of Engineering , Rafik Hairri University, Mechref-Chouf2010, Lebanon. (correspondence e-mail: berjaouisw@rhu.edu.lb)

## I. INTRODUCTION

ROADS form a basis for people transportation and joining between different places. The size of roads varies based on their functionality. For instance, highways are large enough to contain many lanes designed for massive traffic. However, roads inside towns are constructed to be smaller and made up of one or two lanes. Roads are vital in people's daily life, so periodic maintenance shall be made to keep them functional and safe. The many roads that exist within a given country make it difficult to have a continuous assessment of roads; therefore, one can't predict the formation of potholes. Pavement distress is the main cause of defects of roads. Pavement distress can be classified into three classes [1]: Pavement distortion (shoving, corrugation, and rutting), fracture (fatiguing, spalling, and cracking), and disintegration (raveling and stripping).

This work focuses on potholes which are considered the worst pavement distress, and their creation is unpredictable. The main reason behind such distortions can be related to a combination of environmental conditions and traffic pavement stresses.

Potholes are a worldwide problem as they cost governments and citizens billions of dollars yearly [2, 3]. 1.25 million people die each year because of road traffic accidents, 34% of which are related to road potholes [4].

Pothole detection can be categorized into three approaches [5]: Vibration-technique approach [6, 7, 8], 3D reconstruction-technique approach (with laser scanner method, stereo vision method, and Kinect sensor method) [9, 10, 11], and Vision-technique approach [12, 13, 14]. Table 1 summarizes and compares different pothole detection approaches based on technology used, response and sense time, processing, cost, pothole characterization, and accuracy of detection [15]. Traditionally, a group of employees was performing the pothole detection through reviewing recorded digital videos captured from roads. This method is costly and time-consuming [16].

The edge detection method has been used in [17]. The authors used Sobel operators on original images and the vector of edges is then detected using the Fuzzy Inference System (FIS). The usage of FIS type 1 remarkably improves results with a neural network. FIS type 2 improved the training cost of neural network. [18] provides a deep review and comparison of the existing pothole detection methods. The vibration method, which is based on an accelerometer sensor, can't predict potholes ahead of time. This is because the vehicle should pass over the pothole for detection. This method cannot differentiate between potholes and other artifacts on the road like bridge joints and road reflectors. Laser scanning systems are classified among the 3D reconstruction methods. This system can detect potholes in real-time. However, such a type is costly and has a short range of detection. Likewise, stereo vision methods are used in pothole detection. The drawback of this method is the high computational effort for pavement surface reconstruction, vibration sensitivity, and the need for perfect alignment of cameras. Kinect sensor, based on infrared technology, is developed by Microsoft. The sensor is regarded to be costly and necessitates a close distance to the pothole. Finally, the vision-based approaches that make use of monocular camera along with some object detection algorithms showed an improvement in real-time pothole detection.

TABLE I
POTHOLE DETECTION APPROACHES

| | Vision-based | Vibration-based | Laser-based | Stereo imaging |
|---|---|---|---|---|
| Device used | Camera | Accelerometer | Laser | Cameras |
| Technology used | 2D imaging | Force and rotation and orientation | 3D reconstruction of the image using light reflection | 3D reconstruction using multiple cameras |
| Response time | High | Low | Low | High |
| Sensing time | While approaching the pothole | While going through a pothole | While approaching the pothole | While approaching the pothole |
| Processing | Complex image processing algorithms | Readings are directly used | Collection of 3D point cloud with their elevations. | A complex process of 3D image construction by combining image from different camera perceptic |
| Cost | High because of delicate parts like lens | Low | High | High |
| Characterization of pothole | Based on size | Based on vibrations | Based on 3D image constructed | Based on 3D image constructed |
| Detection at night time | Difficult due to poor lighting | Can detect | Can detect | Difficult due to poor lighting |
| Accuracy | Depends on the algorithm used | High | High | Depends on the alignment of cameras and algorithms used |

A pothole detection system made up of Raspberry Pi, camera, 3G mode, and Micro SD card was proposed to be affordable and simple [19]. The authors made use of OpenCV and the system is installed in a stationary manner to report potholes in real-time. The method is based on collecting different frames and convert the picture into a blurring grayscale image. Finally, and after applying some morphological functions, edge detection is employed to define the contours which are fed to Hough transform. Following those few steps, the author was able to detect the presence of a pothole. The system sends an automatic email containing the output image and its location to the transportation department in case a pothole is detected. Another research was able to detect potholes using morphological methods on videos [20]. The authors used digital image processing methods to find the shapes of different objects. The results achieved are recoded to be 93% for accuracy, 93% for precision, and 100% for recall. In another work [21], the authors claimed that a stereo vision system can work in real-time by generating a disparity map for the road with high accuracy and by generating a quadratic fitting in the world coordinate system point cloud. The paper presented the pothole detection successfully without showing any metric of evaluation. An algorithm that can work for the black-box camera is proposed in [22] to create a specialized vision-based system for pothole detection. They claimed that their solution is a low-cost solution and can work in real-time [23, 24]. The authors' work is divided into 3 stages, pre-processing, candidate extraction, and cascade detector. In the first step, the authors extracted the dark areas from a grayscale image, then the lanes of the road are used to find the vanishing point to create virtual lanes, and finally, the pothole region is extracted using some threshold values. They achieved 71% for recall and 88% for precision. This work is limited due to its total dependence on digital image processing techniques. The limitation is in distinguishing potholes from other objects in front of the camera.

Moreover, one paper made use of removing unwanted information from the images' border, like foliage and plants on the border of the street which can lead to wrong readings [25]. Then the convex hull algorithm is used to build a convex contour over the furthest points for different points of interest. The detection of the pothole is based on differences in the color where a large dark shadow area is considered as a pothole. The performance of the system was tested on 53 images with 97 potholes. The results were acceptable with 81.8% precision and 74.4% recall. This technique was able to reject vehicles that appear in the image by an 80% success rate. Furthermore, it was possible to increase the accuracy by more dilation; but this can affect negatively and leads to loss of pothole detection. The system works from a distance of 20 meters down to 2 meters. The computational time of the algorithms is recorded to be 0.148 sec to extract the road from the image and 0.037 sec to find potholes from the image frame.

A new technique based on a thermal IR sensor or camera for pothole detection is proposed in [15]. The method is based on the temperature difference of objects from the surrounding. The captured thermal images of the road are fed into a Convolutional Neural Network (CNN). The paper tested different models of CNN: Self-built CNN models and CNN-based ResNet models. The system with ResNet152 achieved an accuracy of 97.08%. Another deep learning algorithm is used to detect potholes [26]. 900 images with potholes were used to train Fast-RCNN. After detecting a pothole, the image of the pothole is saved along with the pothole GPS and the time of its detection. The collected data are saved in a cloud server. The authors claimed to achieve an average precision exceeding 93%. However, the images of the potholes were taken from a close view of the pavement. A dashboard camera along with a CNN is used also for pothole detection in another work [27]. The neural network consisted of four convolutional and pooling layers and one fully connected layer. The team captured images from different places and had variant conditions, like dry, wet, and shady potholes. The images collected were resized to 200x200 pixels and cropped to remove unwanted parts of the image, they only kept the pothole. After data augmentation is used to get a larger dataset for training with 13244 images, 3250 images for validation, and 500 images for testing. The system achieved 99.8% accuracy, 100% precision, and 99.6% recall. The authors highlighted that their model performed better than the Support Vector Machine (SVM). Furthermore, [28] presented a research on detecting different road damages. This paper studied variant types of road distortion such as white line blur, crosswalk blur, rutting, bump, pothole, alligator crack, construction joint part, wheel mark part, etc. The dataset collected is made up of 9,053 road damage images. The authors adopted object detection methods using CNN in training their system and detecting road distortions. The authors also tested their object detection on smartphones. They achieved 71% and 77% for recall and precision respectively.

A combination of vision and vibration methods for pothole detection has been proposed in [29]. The authors used an accelerometer and camera of mobile phone for this task. Based on SSD with MobileNet, the system was able to detect potholes with 55% accuracy for vibration-based method, and 60% for vision-based method. Although the system was able to detect the pothole in real-time, this would be only done at a close distance to the pothole. A further work proposed a system to detect transverse cracks, longitudinal cracks, and potholes [30]. The authors used mobile phones to get images for those road damages, then sent the data with their GPS location to an online server to evaluate the road damage severity. YOLOv2-tiny is recommended among the tested different deep learning object detectors. Another research made use of digital image processing with spectral clustering to find potholes [31]. The system achieved 81% accuracy, but with the images being cropped around the pothole. An evaluation of the vision-based system using YOLOv2, YOLOv3, and YOLOv3 Tiny has been conducted [32]. The system is trained on 80% of 1500 images of Indian roads. The paper recommended the YOLOv3 Tiny with 72.12 mean Average Precision at 25% Intersection Over Union threshold, 76% for precision, and 40% for recall. However, the speed of processing is not presented. YOLO with ResNet-50 is trained to detect potholes and bumps based on 3399 images for normal roads, 1337 images for potholes, and 547 images for bump [33]. 80% of the dataset was used for training and the rest for validation. The system achieved 88.9% true positive detections. Another research used 448 images, 50% of the images for the training, to train YOLO neural network [34]. YOLOv3, YOLOv3 Tiny, and YOLOv3 SPP achieved mean average precision of 83.43%, 79.33%, and 88.93% respectively, and accuracy was 64.45%, 53.26%, and 72.10% respectively. The database was built by scanning the road from the top view, using a camera mounted on the rear side of an SUV car. The processing of each image is 4ms. On the other side, the gyroscope and accelerometer of a smartphone are used for pothole detection [35]. Inception V3 as "Transfer Learning", is used for feature extraction in CNN. The authors used 70% of the data collected to train CNN and 30% to test it. Their results exhibited a 100% correct classification rate after testing the trained model.

In [36] the accelerometer data collected from mobile phones are normalized by Euler angle computation which is fed to a combination algorithm of Z-THRESH and G-ZERO approaches. Then, spatial interpolation is used to locate the pothole. Results revealed a 100 % accuracy in detecting potholes without false positives. Another work presented a real-time system for inspecting and detecting road distresses [37]. The system used a high-speed 3D transverse scanning method. Structured light triangulation formed the base of the characterization and dynamic generation of the 3D pavement profile. The detection system is made up of a GigE digital camera and an infrared laser (810 nm) line projector. The system is mounted on the rear side of the vehicle. To make the laser stripe covers a full lane of pavement transversely, an 80° fan angle laser projector has been used. The camera catches continuous images for the lines of the laser to compute 3D transverse profile. Based on the triangulation principle, the elevation of a specific point can be found. Another work employed laser imaging for pavement distress inspection [38]. Several features are then captured including the total number of distress tiles and the depth index which are given to a three-layer neural network for classifying the type of the crack and to estimate its severity.

Another research achieved a high recognition rate of potholes based on texture measures from the histogram as features to train the Support Vector Machine (SVM) [39].

To conclude, the vibration method is a low cost method and can evaluate pothole severity. However, this method could be harmful to the car and it can't differentiate between potholes and other artifacts on the road. In the 3D reconstruction technique, the laser scanner method can estimate pothole size and its severity, but this system is expensive and has a short range of detection. The stereo vision method has an average price and can estimate pothole size, but it can't evaluate pothole severity. The alignment of cameras must be perfect and has a short range of detection. The Kinect method is still a new way in detecting potholes and in evaluating the pothole size and severity, but this method can't operate under the sunlight and it has a short range of detection. For vision technique, it's cheaper than most of the previously mentioned methods and can detect potholes time ahead. It can estimate the pothole size but can't evaluate pothole severity.

To build an efficient pothole detection model, a pothole dataset is created for Lebanese and other countries' roads. The model of detection is improved to work under real-life conditions. The system has high processing capability and achieved high frame/second (FPS) to detect potholes time ahead. Our system is robust with acceptable precision and sensitivity/recall.

This work can be used to report potholes from roads in real-time to responsible agencies, increase the safety of drivers by helping them detect potholes time ahead, and the proposed system could improve the performance of self-driving cars to ensure safe trips for passengers in the future.

## II. METHODS

### A. Applied Object Detection

Object detection process served under different variations in poses, occlusions, viewpoints, and lighting conditions of the input data [40]. The pipeline object detection model can be divided into three parts: informative region selection, feature extraction, and classification [40]. At the informative region selection, the objects that may present in an image have different sizes, aspect ratios, or specified positions. A multi-scale sliding window is, therefore, used to scan the whole image. Due to limitless redundant candidate windows, its computation is viewed to be expensive. On the other hand, unsatisfactory regions can be detected if only a fixed number of sliding window templates are applied [40]. At the feature extraction, one should extract visual features with robust and semantic representation to recognize different objects. Designing a robust feature descriptor to match and detect all kinds of objects is somehow difficult. This is due to the diversity of illumination conditions, appearances, and backgrounds [40]. At the classification, the Deformable Part-based Model (DPM), Support Vector Machine (SVM), and AdaBoost are used as classifiers to distinguish object classes. They also make the representations more hierarchical, semantic, and informative for visual recognition [40]. Object detection has many architectures that can be used for model training and can be divided into two types. The first type is a classification that is based on selecting the region of interest and then classifying each region into various object classes. Examples of these architectures are R-CNN, SPP-net, Fast R-CNN, Faster R-CNN, Mask R-CNN, etc. The second type is regression at which is based on adopting a unified architecture to achieve results directly. The best two known examples from this group are the SSD (Single Shot Multibox Detector) and YOLO (You Only Look Once) family algorithms [41].

### B. Datasets

In this study two datasets are used: the first database is available online [42] and made up of 431 different images with potholes. The second dataset is a combination of images collected from Lebanese roads (with 344 images with potholes) and 312 images were collected from different sources on the internet. Most of them are from videos that were recorded via dashboard camera for people driving their car. As a result, 1087 different images with more than 2000 potholes are used in this study. **Error! Reference source not found.** illustrates a sample of an image captured by a smartphone from Lebanese roads.

### C. Architectures and Techniques

#### 1) TensorFlow Framework

TensorFlow is a machine learning framework with an end-to-end open source. It has flexible, comprehensive tools, libraries, and community resources that allow users and researchers to push the state-of-the-art in ML develop, and deploy ML powered applications [43]. There are more frameworks other than TensorFlow, such as Caffe, Caffe2, PyTorch, and more.

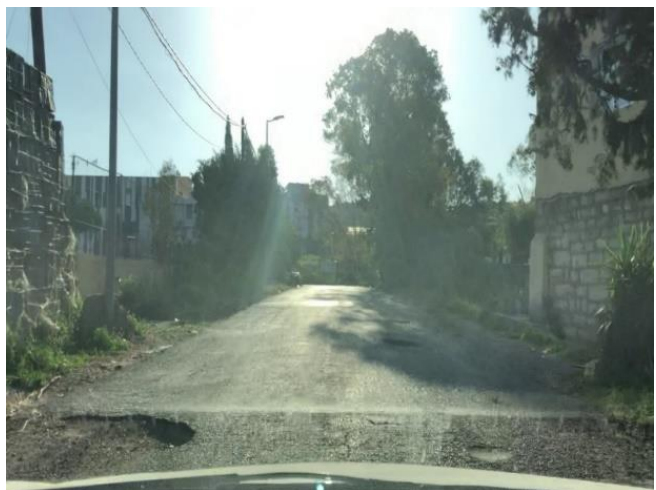

Fig. 1. Sample of an image from the used dataset

But TensorFlow is one of the most important frameworks and has a large community of support, also it builds and trains a model without sacrificing speed and/or performance.

#### 2) Darknet Framework

Darknet is an open-source framework like TensorFlow and written in C/CUDA, and it is used to train neural networks and serves as the basis for YOLO.

#### 3) Single Shot Multibox Detector (SSD)

SSD is an object detection model, it was published in 2016 by researchers from Google. It uses a single deep neural network combining feature extraction and regional proposals [44].

#### 4) You Only Look Once Detector Version 3 (YOLOv3)

YOLO can produce decent boxes for detected objects [45]. YOLO is an abbreviation of "You Only Look Once", and its name is inherited from the fact that: you only look once at the input image to predict what objects present in addition to their position in an image. Unlike other models which require a scan of an image several times. YOLO is an increment work started in 2015 with YOLOv1, YOLOv2 in 2016, YOLOv3 in 2018, and YOLOv4 in 2020. **Error! Reference source not found.** plots the accuracy vs speed of YOLOv3 against RetinaNet-50 and RetinaNet-101. Clearly, YOLOv3 brings a significant advantage over other detection models in terms of speed [45].

#### 5) You Only Look Once Detector Version 4 (YOLOv4)

YOLOv3 is known to be a strong detector that can produce decent boxes for objects detected in real-time. YOLOv4 comes as an improvement of YOLOv3. [46] claims that YOLOv4 is 10% better than the YOLOv3 in AP, and 12% for the speed. YOLOv4 is a recently released architecture by April 2020.

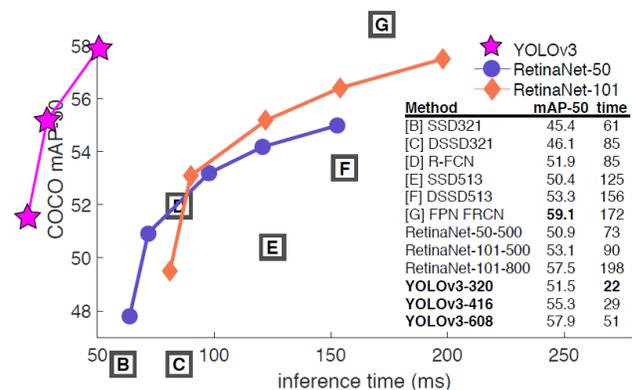

Fig. 2. Detection Models Performance [45]

#### 6) Google Colaboratory

Colaboratory (Colab) is made by Google Research. Colab is a Linux machine and its interface is based on the Jupyter notebook service and it doesn't need setup. It provides free access to decent computing resources like Graphical Processing Unit (GPU) [47]. Accordingly, it is used to develop deep learning applications using popular libraries such as Keras, TensorFlow, PyTorch, and OpenCV. This virtual machine runs on a 2core CPU, 12GB of RAM and can be upgraded to 25GB if needed for free. For the GPUs, one will get a random GPU when running the notebook. The GPUs that can be accessed are Nvidia Tesla K80s, T4s, P4s, and P100s [47].

### D. *Training on Colab and Preparation*

In this work, TensorFlow is used as a framework and SSD as architecture to build a real-time model that can detect potholes. [48] Serves as a good reference for implementing SSD using TensorFlow. The images of the first dataset are annotated before loading them into the framework. Image annotation is the process of labeling the objects to be detected. The software used to label the data is called "LabelImg", an abbreviation of Label Image [49]. The labeling process can be done by drawing a bounding box around the potholes. The position of each pothole is then converted into a text file with the coordinates of the bounding box having the following format (Xmin; Ymin; Xmax; Ymax). The same procedure is used to prepare the YOLOv3 and YOLOv4 training, but the framework used is Darknet.

To start training, a new notebook was created on Colab and the hardware accelerator option is enabled by selecting the GPU option. GPUs are known to be much faster than Central Processing Unit (CPU) for our deep learning application. There are libraries to optimize the use of GPUs in deep learning as the NVIDIA CUDA Deep Neural Network Library (cuDNN), known as a GPU-accelerated library for deep learning [50].

## III. EXPERIMENTS

Given that all preparations of data and training are done on COLAB with a GPU hardware accelerator, the following experimentations are done:

### A. *SSD-TensorFlow for Pothole Detection on Colab*

SSD is considered a fast object detection architecture and TensorFlow is its framework. In this experiment, the first dataset is divided into 391 images for training and 40 images for testing. The image resolution during the training phase was 1220×920. Equations (1), (2), and (3) are used for classification loss, localization loss, and total loss respectively [51].

$$L_{conf}(x,c) = - \sum_{i \in Pos}^{N} x_{ij}^p \log(\hat{c}_i^p) - |\sum_{i \in Neg} \log(\hat{c}_i^0) \qquad (1)$$

Where $\hat{c}_i^p = \frac{\exp(c_i^p)}{\sum_p \exp(c_i^p)}$

$$L_{loc}(x,l,g) = \sum_{i \in Pos}^{N} \sum_{m \in \{cx,cy,w,h\}} x_{ij}^k smooth_{L1}(l_i^m - \hat{g}_j^m) \qquad (2)$$

$$L_{(x,c,l,g)} = \frac{1}{N}(L_{conf}(x,c) + \propto L_{loc}(x,l,g)) \qquad (3)$$

The localization loss is a Smooth L1 loss between the ground truth box (g) and the predicted box (l) parameters [52]. (cx; cy) are the offset center of the default bounding box (d), height (h), and width (w). The confidence loss is the loss over multiple confidence classes(c). During cross-validation, the weight term α is set to one and the N term is the number of matched default boxes [51].

The training took around 15 hours and ran for 144.7k steps. A 73% Positive predictive value and a recall of 37.5% with 32.5% mAP have been achieved. Accordingly, the results were not satisfying also the maximum frames per second (FPS) achieved is no more than 10 FPS. The total loss, a combination of both classification and localization losses, is shown in **Error! Reference source not found.**.

### B. *YOLOv3- Darknet53 for Pothole Detection on Colab*

The steps of YOLO preparation are different than SSD-

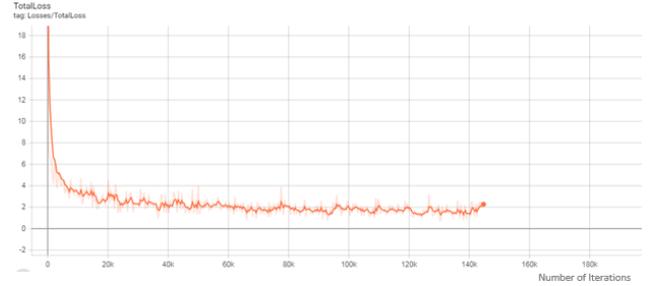

Fig. 3. Total Loss

TensorFlow. For instance, the format of the bounding box in YOLO is given as (class, Xcenter of Bounding box, Ycenter of Bounding box, width of the bounding box, height of the bounding box). YOLO is also known as a fast object detection architecture and darknet is its framework. Six different training scenarios are conducted as shown in **Error! Reference source not found.**.

The trainings scenarios in **Error! Reference source not found.** would actually help in finding the best configurations

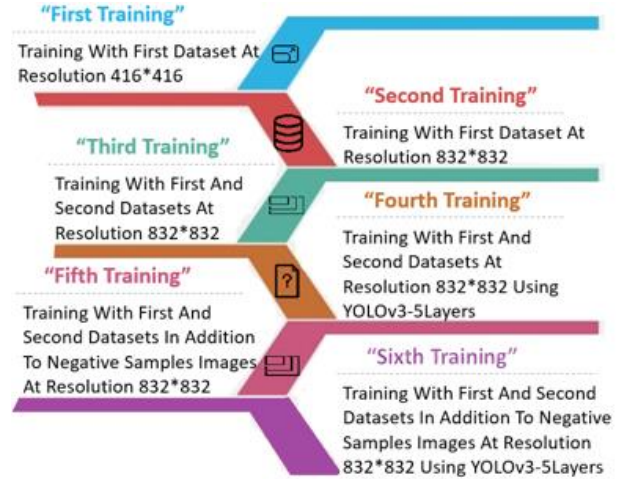

Fig. 4. Different Training Scenarios

for YOLOv3 detector in our pothole detection application. To check the performance and robustness of the model at each scenario, four different configurations of confidence and IoU thresholds were used at every 250 iterations:
- Test 1: confidence threshold 25% and IoU threshold 50%
- Test 2: confidence threshold 25% and IoU threshold 10%

- Test 3: confidence threshold 25% and IoU threshold 1%
- Test 4: confidence threshold 50% and IoU threshold 1%

The time requested by each training to achieve 5000 iterations is summarized below:

- First Training: 10.5 hours
- Second Training: 20.75 hours
- Third Training: 21.25 hours
- Fourth Training: 60 hours
- Fifth Training: 26 hours
- Sixth training: 65 hours

The True Positive Rate (TPR), False Positive Rate (FPR) and Mean Average Precision (mAP) are measured at each training scenario over the four different tests. The first training demonstrated a decrease in IoU threshold during test2 and test 3 compared to other tests which increased the true detections as shown in **Error! Reference source not found.**a. Test2 served a better TPR and minimal false detections in comparison to the rest of 4 tests. Test 3 has been selected For the second training as shown in **Error! Reference source not found.**b, the third training is shown in **Error! Reference source not found.**c, the fourth training is shown in **Error! Reference source not found.**d, the fifth training is shown in **Error! Reference source not found.**e, and the sixth training is shown in **Error! Reference source not found.**f. Test3 gave the highest mAP and minimal false detections among other tests. The key concept in the proposed system is to give more attention to the true detections achieved rather than focusing on the false detection with an acceptable range. YOLOv3-Darknet53 Pothole detection speed (FPS) is highly dependent on the GPU used and the resolution of the images. The GPU used is NVIDIA Tesla P100-PCIE as it is viewed as the best GPU on COLAB. When the resolution is set to 416× 416, a 50 FPS is achieved compared to 19 FPS when the resolution was set to 832×832.

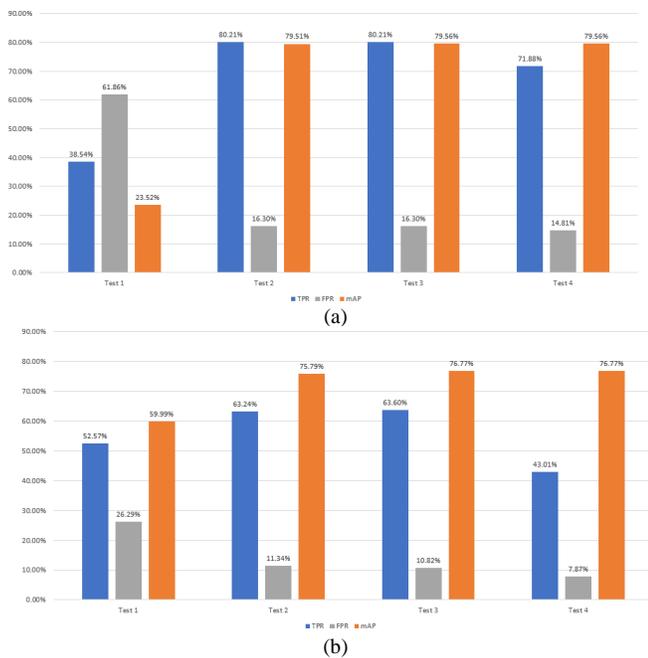

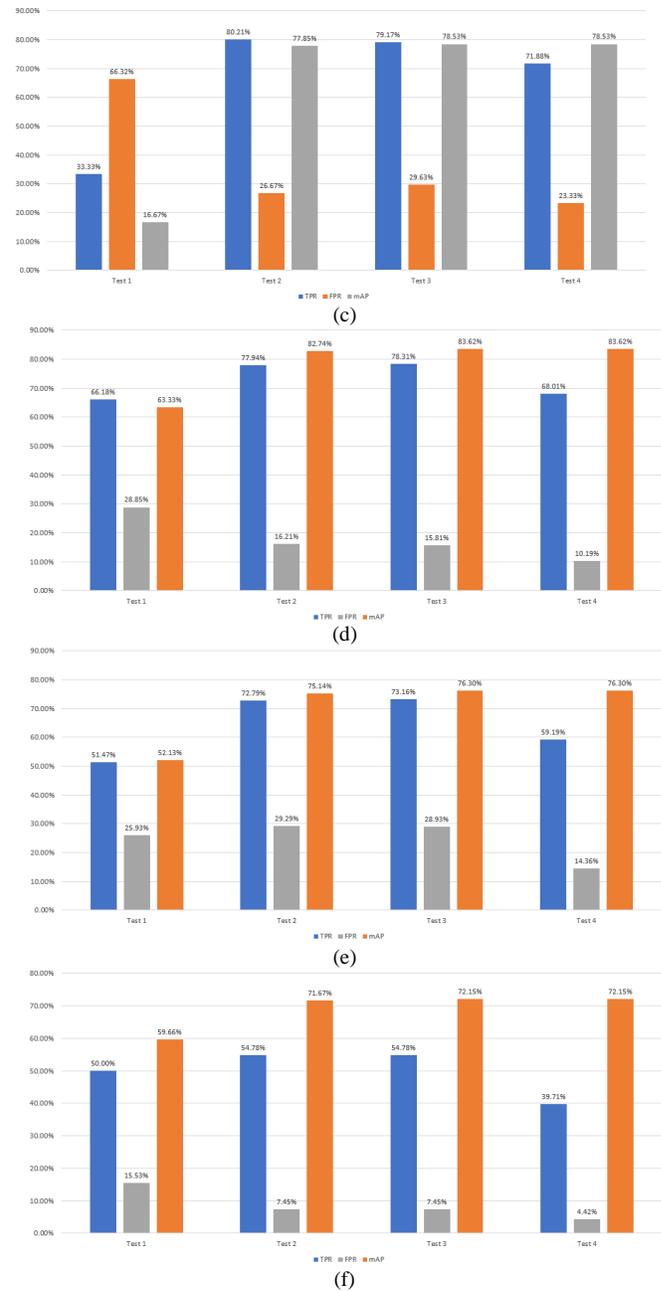

Fig. 5. (a) First Training (after 3000 iterations), (b) Second Training (after 3000 iterations), (c) Third Training (after 3250 iterations), (d) Fourth Training (after 2750 iterations), (e) Fifth Training (after 4000 iterations), (f) Sixth Training (after 3500)

### C. YOLOv4- CSPDarknet53 For Pothole Detection on Colab

YOLOv4 is a new architecture released in April 2020 and this work is the first of its kind to use YOLOv4 for pothole detection applications. Two pieces of training were done to evaluate YOLOv4 in potholes detection. The first training contains the first and second datasets and trained like the YOLOv3, and the second training contains an additional 941 negative images. Negative images are images that don't contain

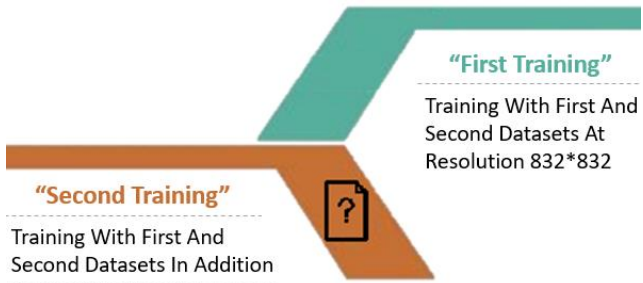

Fig. 6. The two training Scenarios on YOLOv4-CSPDarknet53

## IV. Results and Discussion

Compared results of YOLOv3 and YOLOv4 can be summarized in **Error! Reference source not found.**.

SDD is not included since SDD exhibited inferior performance in comparison with YOLO. YOLOv4 proved to be a robust object detector architecture. It can work with real-life conditions and at a high speed of processing. The proposed system achieved high recall 81%, high precision 85%, and 85.39% mAP. Moreover, this pothole detector attained a processing speed up to 21 FPS using Colab GPU, NVIDIA Tesla P100-PCIE. This system can work with raw data collected from dashboard Cameras for roads. In other words, our system can work without the need to cropping and deleting non-pothole data from the input images. Additionally, the proposed pothole detector can detect a pothole from a distance potholes. The two training scenarios were used to find the best configuration for YOLOv4 pothole detection as illustrated in **Error! Reference source not found.**:

As in YOLOv3, the performance and robustness of the model have been tested every 250 iterations weight under the four different test configurations of confidence and IoU. The training time took 22 hours in the first training and 25 hours in the second training to achieve 5000 iterations.

The first training showed better performance with higher

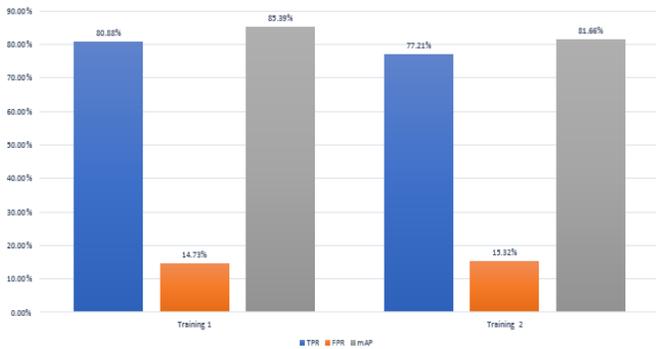

Fig. 7. YOLOv4 First and Second Trainings Performance Comparison

mAP and lower FPR than the second training scenario as in Fig.

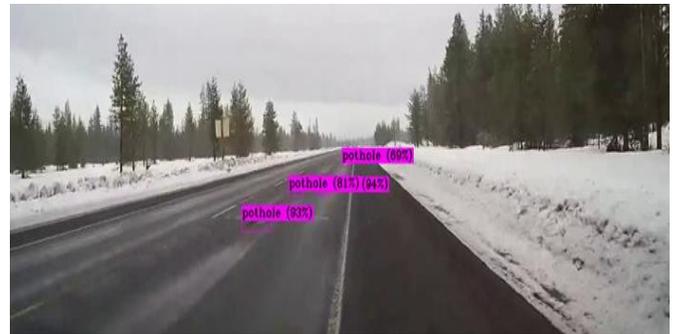

Fig. 9. Testing YOLOv4 on real life scenario

reaching 100 meters. Detection made ahead of time can help drivers in avoiding potholes. A sample of the performance of YOLOv4 training is shown in **Error! Reference source not found.**.

A video of the proposed system performance can be found on (https://youtu.be/UjAHRV0oHbE). To put our system into practice, the pre-trained model has been tested on Lebanese

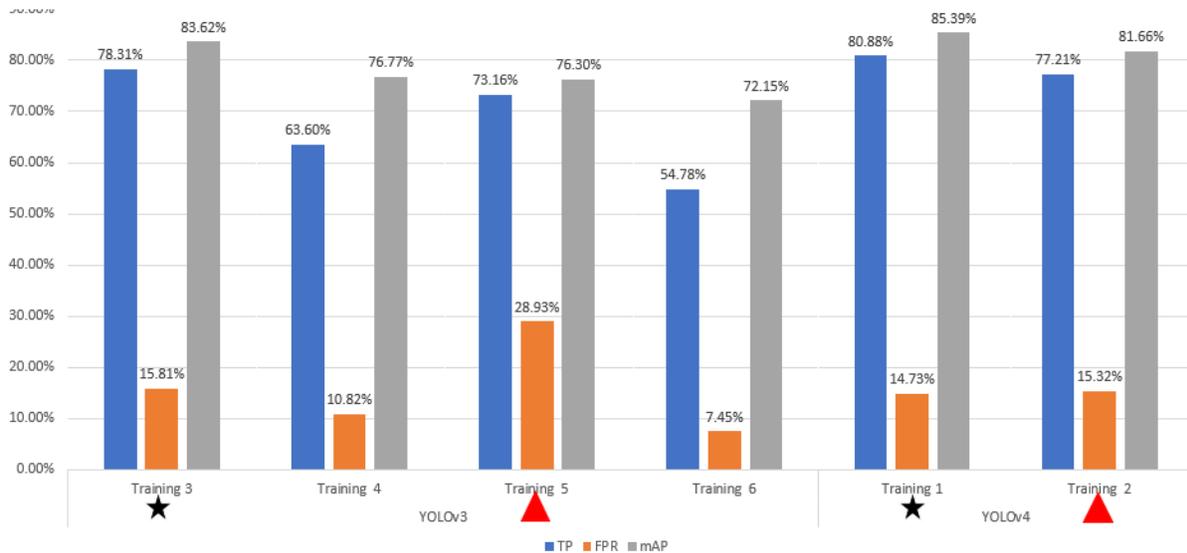

7. Configurations of test 3 yielded those results.

road using the smartphone iPhone X with a 12-megapixel wide-angle camera. The camera has a focal ratio of f/1.8. The tested system is shown on (https://youtu.be/-RH_fRhQsvs).

Fig. 10 displays a comparison between our pothole detection models and other models discussed in the literature review. Some systems have a good performance close to our system like the "Pothole Detection System Using A Black-box Camera" [22]. However, the system has a short range of pothole detections in comparison with the tested YOLOv3 and YOLOv4 models. Furthermore, the system needs to have a clear view of the car path which is not practical in real life. "Deep Learning-Based Detection of Potholes in Indian Roads Using YOLO" [32] has a good range of detection, but the evaluation metrics were better in our model.

current system was able to differentiate between manholes and potholes in most of the cases. Moreover, we aim to deploy our system into several cars to analyze the road condition in a live and real-time manner by adding GPS to get the coordinates of the pothole for maintenance.

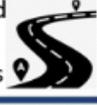

Fig. 10. Comparison with cutting-edge techniques on pothole detection application

## V. Conclusion

The potholes detector using SSD-TensorFlow achieved 73% and 37.5% for the precision and recall respectively, and 32.5% for the mAP. YOLOv3 achieved high recall 78%, high precision 84%, and 83.62% mAP. YOLOv4 achieved high recall 81%, high precision 85%, and 85.39% mAP. The speed of processing for SSD was low and can't be used for real-time applications. YOLOv3 and YOLOv4 have a speed of processing of around 20 FPS and this can be considered high enough for our real-time application. Therefore, our potholes detector using YOLOv4 can be considered as a robust and real-time system that can be used in real-life scenarios.

Our future work will include a larger dataset, more than 2000 images, for training and it has more images for potholes from different roads, with several severities and different lighting and weather conditions. Alexey AB [53] and many other researchers emphasized using a large dataset with more than 2000 images for training to get a robust object detector that can work under any conditions and circumstances. Moreover, in our future work, we will include manholes in the training of our system. Manholes and potholes have common features and this important improvement to be done into our current system to differentiate between manholes as a pothole. Knowing that the